\documentclass[article, 10 pt, conference]{ieeeconf}  
\IEEEoverridecommandlockouts                              

\UseRawInputEncoding                                    

\usepackage[
top    = 0.75in,
bottom = 0.75in,
left   = 0.75in,
right  = 0.75in]{geometry}
\usepackage{cite}
\usepackage[skip=10pt plus1pt, indent=35pt]{parskip}
\usepackage{placeins}
\usepackage{textcomp}
\usepackage[dvipsnames]{xcolor}
\usepackage{booktabs}
\usepackage{multirow}
\usepackage{siunitx}
\usepackage{float}
\usepackage{paralist}
\usepackage{svg}
\usepackage{array}
\usepackage{subcaption}
\usepackage{xspace}
\usepackage{scrextend}
\usepackage{graphicx} 
\usepackage[export]{adjustbox}
\usepackage{caption}
\usepackage{tabularx}
\usepackage{listings}
\usepackage{wrapfig}
\usepackage{calligra}
\usepackage{balance}
\newcolumntype{A}{ >{\centering\arraybackslash} m{4cm} }
\newcolumntype{B}{ >{\centering\arraybackslash} m{1cm} }

\usepackage{amsmath} 
\usepackage{amssymb}  
\usepackage{amsfonts}
\usepackage{authblk}
\makeatletter
\let\NAT@parse\undefined
\makeatother
\usepackage[colorlinks=true]{hyperref}  
\def\BibTeX{{\rm B\kern-.05em{\sc i\kern-.025em b}\kern-.08em
    T\kern-.1667em\lower.7ex\hbox{E}\kern-.125emX}}

\title{\LARGE \bf
Anticipate \& Act : Integrating LLMs and Classical Planning for Efficient Task Execution in Household Environments $^\dagger$

}

\author{Raghav Arora$^{1*}$, Shivam Singh$^{1*}$, Karthik Swaminathan$^{1}$, Ahana Datta$^{1}$, Snehasis Banerjee$^{1,2}$, \\ Brojeshwar Bhowmick$^2$, Krishna Murthy Jatavallabhula$^{3}$, Mohan Sridharan$^4$, Madhava Krishna$^1$

\thanks{* Denotes equal contribution}
\thanks{$^{1}$ Robotics Research Center, IIIT Hyderabad, India}
\thanks{$^{2}$ TCS Research, Tata Consultancy Services, India}
\thanks{$^{3}$ CSAIL, Massachusetts Institute of Technology, USA}
\thanks{$^{4}$ IPAB, University of Edinburgh, UK}
\thanks{$^\dagger$ This work was funded by TCS Research, India}
}

\begin{document}

\maketitle
\thispagestyle{empty}
\pagestyle{empty}

\begin{abstract}
  Assistive agents performing household tasks such as making the bed or cooking breakfast often compute and execute actions that accomplish one task at a time. However, efficiency can be improved by anticipating upcoming tasks and computing an action sequence that jointly achieves these tasks. State-of-the-art methods for task anticipation use data-driven deep networks and Large Language Models (LLMs), but they do so at the level of high-level tasks and/or require many training examples. Our framework leverages the generic knowledge of LLMs through a small number of prompts to perform high-level task anticipation, using the anticipated tasks as goals in a classical planning system to compute a sequence of finer-granularity actions that jointly achieve these goals. We ground and evaluate our framework's abilities in realistic scenarios in the \emph{VirtualHome} environment and demonstrate a $31\%$ reduction in execution time compared with a system that does not consider upcoming tasks. 
  Website: \href{https://raraghavarora.github.io/ahsoka}{https://raraghavarora.github.io/ahsoka}
  
\end{abstract}
\vspace{-1em}

\section{Introduction}
\begin{figure}[tb]
\centering
\captionsetup{font=scriptsize}
\setlength{\belowcaptionskip}{-10pt}
\includegraphics[width=0.4\textwidth]{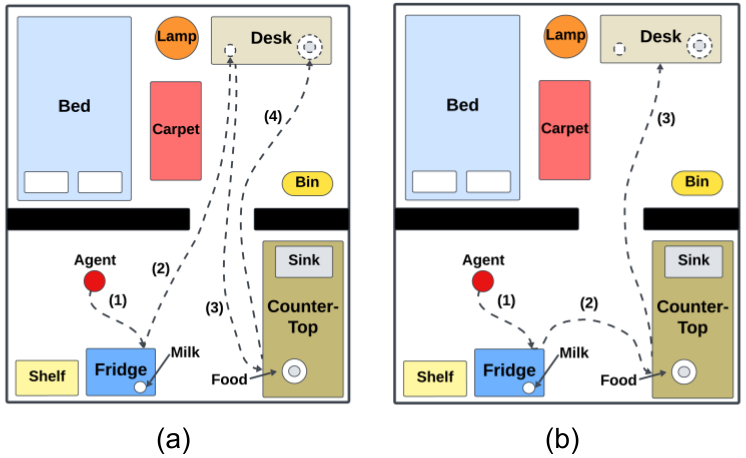}
\caption{Anticipation example: (a) Agent individually moves the milk and then the food to the desk; (b) Agent anticipates that milk needs to be served after food, jointly moving them to eliminate an extra trip.}
\vspace{-6pt}
\label{fig:teaser}
\end{figure}

Consider an agent assisting humans with daily living tasks in a home. 
In the scenario in Fig.~\ref{fig:teaser}, these tasks include making the bed and cooking breakfast, with each task requiring the agent to compute and execute a sequence of finer-granularity actions, e.g., fetch the relevant ingredients to cook breakfast. Since the list of tasks can change based on resource constraints or the human's schedule, the agent is usually asked to complete one task at a time. However, the agent can be more efficient if, similar to a human, it anticipates and prepares for upcoming tasks while computing a plan of finer-granularity actions, e.g., it can fetch the ingredients for breakfast when it fetches milk to make coffee.\\
\indent State-of-the-art methods for estimating future tasks or their costs formulate them as learning problems and use data-driven deep networks~\cite{8460924,dhakal2023anticipatory}. There is also work on using Large Language Models (LLMs) for task planning~\cite{pmlr-v205-huang23c,ding2023task,lin2023text2motion}. However, these methods predict sequences of high-level tasks or require a large labeled training dataset to compute a sequence of fine-grained actions. 
They also make it difficult to leverage domain knowledge, adapt to environmental changes, or to understand the decisions made.\\
\indent We pose high-level task anticipation and finer-granularity action execution as a combined prediction and planning problem. Our framework seeks to leverage the complementary strengths of data-driven estimation based on generic prior knowledge of household tasks, and planning based on domain-specific action theories. Specifically, our framework: 
\\\indent 1. Leverages the generic knowledge encoded in LLMs using a small number of prompts describing potential sequences of high-level household tasks, in order to predict subsequent tasks given partial sequences of tasks.
\\\indent 2. Uses an action language to encode finer-granularity domain-specific knowledge of household tasks in the form of domain and agent attributes, agent actions, axioms governing change, and heuristics to guide planning.
\\\indent 3. Adapts a symbolic and heuristic classical planner to consider both the immediate and the anticipated tasks as goals, computing a sequence of actions to jointly minimize the cost of accomplishing these goals. 
\begin{figure*}[!htbp]
\centering
\captionsetup{font=scriptsize}
\setlength{\belowcaptionskip}{-10pt}
\includegraphics[height=0.4\textwidth,width=0.95\textwidth]{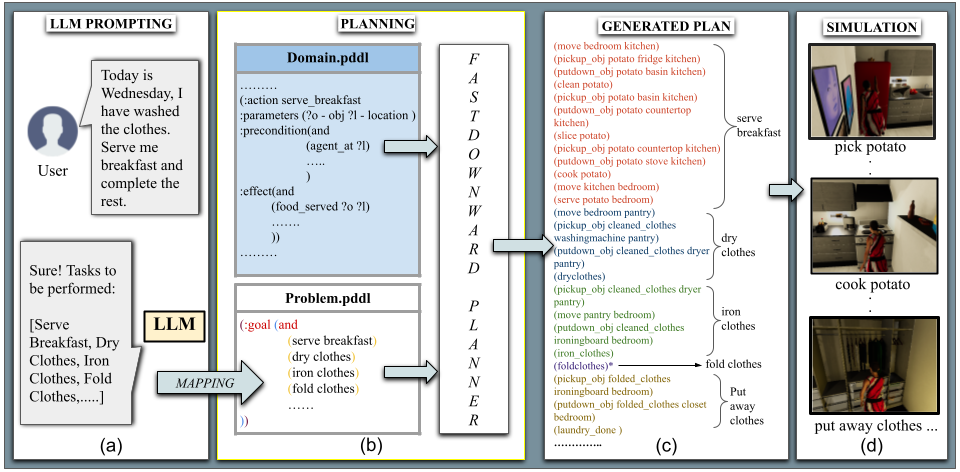}
\caption{Our framework's pipeline: (a) user inputs prompts with sequences of household tasks to an LLM, which then predicts high-level tasks over a time horizon; (b) the sequence of tasks is mapped to a joint goal state in a finer-granularity domain description in an action language (PDDL); (c) a heuristic planner (FD) uses this description to jointly compute the sequence of actions to be executed to complete all the tasks; and (d) the plan is executed in a realistic simulation environment.\\$^*$Actions corresponding to \texttt{fold clothes} are omitted due to space restrictions.}
\label{fig:pipeline}
\end{figure*}
\\\indent We prompt existing LLMs for high-level task anticipation, use the Planning Domain Definition Language (PDDL)~\cite{pddl} as the action language, and use the Fast Downward (FD) solver~\cite{Helmert_2006} to generate fine-granularity plans for any given task. We evaluate our framework's abilities in \emph{VirtualHome}, a realistic simulation environment~\cite{puig2018virtualhome}, and in complex household scenarios involving multiple tasks, rooms, objects, and actions. We present a $31\%$ reduction in execution time and a $12\%$ reduction in plan length compared to a system that does not anticipate upcoming tasks.
\vspace{5pt}
\section{Related Work}

In this section, we review the related work to motivate our framework for task anticipation and plan generation.

\vspace{-0.75em}
\textbf{LLMs for robotics:} LLMs such as GPT-4 \cite{OpenAI2023GPT4TR}, PaLM~\cite{chowdhery2022palm}, and Llama~\cite{touvron2023llama} are being used to address problems in robotics and AI. This includes generating ``plans" to achieve goals~\cite{huang2022inner,ding2023task,Sharma2021SkillIA}, using descriptions of plans extracted from different sources~\cite{Ahn2022DoAI,huang2022language, Lin2023SwiftSageAG,lin2023text2motion}. Some methods have proposed prompting strategies to validate and improve previously generated plans~\cite{raman2022planning, silver2023generalized}. Other methods have demonstrated that the LLM-based summarization can be used for perception and scene understanding~\cite{wu2023tidybot}, and to generate code for planning and robot manipulation~\cite{silver2023generalized,10161317,huang2023voxposer}. Since LLMs are trained on a large amount of data from the Web to predict the text likely to appear next, researchers have questioned the ability of LLMs to plan (in the classical sense) based on prior domain knowledge~\cite{collins2022structured,valmeekam2023large}. 


\vspace{-0.75em}
\textbf{Task Anticipation:} Knowledge-based and data-driven methods have been developed for task anticipation, with state-of-the-art methods using deep networks~\cite{dhakal2023anticipatory} and LLMs~\cite{zhao2023antgpt}. These methods predict high-level tasks or their costs in simplistic domains, with additional planning required to complete each such task, or require many examples to directly predict a sequence of finer-granularity actions to be executed.
Our framework, on the other hand, leverages the complementary strengths of LLMs and classical planning. It enables an agent to operate in complex environments by anticipating high-level tasks based on limited prompts to an LLM, using these tasks as goals to be achieved jointly by planning a sequence of finer-granularity actions based on domain-specific knowledge. 

\vspace{-0.75em}
\textbf{Integrating LLMs with PDDL:} Given the existing literature on using PDDL to encode prior knowledge for planning~\cite{pddl}, recent papers have emphasized the need for such planning in combination with LLMs in complex domains~\cite{silver2023generalized, silver2022pddl}. LLMs have been used to generate (or translate prior knowledge to) goal states to be achieved by a classical (PDDL-based) planner~\cite{liu2023llmp, xie2023translating}. 
However, research has also indicated that methods based on deep networks and LLMs are not well-suited for multistep, multilevel decision-making (in the classical sense) by reasoning with domain knowledge~\cite{valmeekam2023large}. 
\section{Problem Formulation and Framework}
\vspace{-3pt}
\label{sec:problem}
Consider an assistive agent asked to complete a routine, i.e., a sequence of high-level tasks $\mathcal{R} = \{\tau_1, \tau_2,...\tau_n\}$. 
In a household environment, each $\tau_i$ is one of a set $\mathcal{T}$ of known tasks, e.g., \textit{make the bed} or \textit{make breakfast}, which requires the robot to compute and execute a plan, i.e., a sequence of finer-granularity actions $\{a_1, \ldots, a_{m_i}\}$. For example, to \textit{water the plants}, the agent has to \textit{bring the water hose to the garden}, \textit{connect the hose to the tap}, and \textit{turn the tap on}. 
Since the agent can be assigned different sequences of high-level tasks, it typically considers one task (in a given sequence) at a time, computing a sequence of finer-granularity actions to complete the task at a minimum cost (or time, effort). In doing so, the agent may fail to leverage an opportunity to reduce the cost of completing a subsequent task, e.g., when the agent is fetching milk from the fridge for coffee, it can also get the ingredients for making breakfast. Our framework in Figure~\ref{fig:pipeline} leverages the generic knowledge of LLMs to anticipate one or more high-level tasks given a partial sequence and a limited number of prompts. The anticipated tasks are considered as goals in a classical planner that computes a sequence of actions to jointly achieve these goals. We describe our framework's components below.

\subsection{LLMs for task anticipation}
LLMs like GPT-3~\cite{brown2020language} can be tuned to predict patterns (of tasks) in specific domains. 
In our household domain, the high-level tasks are daily living tasks, e.g., \textit{iron clothes} and \textit{vacuum the house}, that are often performed at specific times and in a specific order, e.g., \textit{wash clothes} and \textit{dry clothes} are completed before \textit{iron clothes}. Our framework uses an LLM to extract these task patterns from a small number of task routines provided as prompts. As described in Section~\ref{sec:expt}, these routines can have $\approx 20$ high-level tasks. Given a partially specified routine, the LLM can predict tasks that the agent is likely to be asked to complete next---see Figure~\ref{fig:pipeline}(a).

Our choice of using the LLMs to model and predict sequences of high-level tasks is motivated by two objectives: (i) exploiting the complementary strengths of generic LLMs and domain-specific knowledge-based planning methods; and (ii) leveraging the capabilities of an LLM with limited examples of routines of interest. As described in Section~\ref{sec:expt}, we explored the use of popular LLMs such as PaLM~\cite{chowdhery2022palm}, GPT-3.5\cite{brown2020language}, and GPT-4\cite{OpenAI2023GPT4TR}. We also explored the effect of using context-specific examples during training or execution. 





\subsection{Action Planning with Anticipated Goals}
The tasks anticipated by the LLM are considered as goals, and our framework uses a classical planner to compute the sequence of finer-granularity actions to be executed to jointly achieve these goals. As stated earlier, the planner uses domain-specific knowledge in the form of a theory of actions; we use the STRIPS~\cite{STRIPS} subset of PDDL~\cite{pddl} enriched with types, negative preconditions, and action costs as the action language to describe this theory. We also focus on goal-based problems and discrete actions with deterministic effects; other action languages can be used to represent durative actions~\cite{gerevini:FI11} or non-determinism~\cite{mohan:JAIR19}.

For any given domain, statements in PDDL describe the \texttt{domain} and the \texttt{problem} to the solved. The domain description $\mathcal{D} = \langle \mathcal{S}, \mathcal{H}\rangle$ comprises a signature $\mathcal{S}$ and a theory $\mathcal{H}$ governing the domain dynamics. The signature $\mathcal{S}$ includes a specification of \textit{types} such as \textit{location}, \textit{object}, \textit{receptacle} and \textit{agent}; \textit{constants} such as \textit{kitchen} and \textit{garden} that are specific instances of the types; and \textit{predicates} that include \textit{fluents}, \textit{statics}, and \textit{actions}. Fluents such as \textit{(agent\_at ?l - location)}, \textit{(obj\_at ?o - obj ?l - location)}, and \textit{(dropped ?o1 - obj ?r - receptacle ?l - location)} represent domain attributes whose values can change over time as a result of actions; \textit{statics} are domain attributes whose values do not change over time; and \textit{actions} such as \textit{move\_agent}, \textit{cook}, \textit{serve}, and \textit{pickup} change the value of relevant fluents. $\mathcal{D}$ also includes a specification of each action in terms of its parameters, preconditions that need to be true for the action to be executed, effects that will be true once the action is executed, and the cost of executing it. For example, Figure~\ref{fig:Action} provides the specification of the \textit{dusting} action; the preconditions are that the agent have a mop in its hand and be in a location with an object that needs to be dusted.

\begin{figure}[tbp] 
\captionsetup{font=scriptsize}
\begin{center}
    \begin{minipage}{0.45\textwidth} 
        \small{
        \begin{lstlisting}[language=Lisp, frame=single, breaklines=true]
(:action dusting
 :parameters(?o - obj ?l - location)
 :precondition(and(In_hand DustMop)  
                  (agent_at ?l)
                  (not(dusted ?o ?l)))
 :effect(and(dusted ?o ?l)(increase (total-cost)10))
)
        \end{lstlisting}
        }
    \end{minipage}
    \setlength{\abovecaptionskip}{0pt}
    \setlength{\belowcaptionskip}{-15pt}
    \caption{Action for \textit{Dusting} the object $o$ at some location $l$.}
    \label{fig:Action}
    \vspace{0.5em}
\end{center}
\end{figure}
\begin{figure}[tb]
\centering
\captionsetup{font=scriptsize}
\vspace{0.3em}
\includegraphics[width=0.48\textwidth,frame]{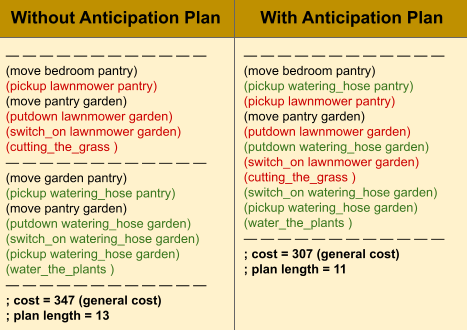}
\caption{Example plans produced with and without considering anticipated tasks.}
\label{fig:comparison_plan}
\vspace{-1em}
\end{figure}
The problem description $\mathcal{P} = \langle\mathcal{O}, \mathcal{I}, \mathcal{G} \rangle$ describes the specific scenario under consideration in terms of the set $\mathcal{O}$ of specific objects, the initial state $\mathcal{I}$ comprising ground literals of the fluents and statics, and a description $\mathcal{G}$ of the goal state in the form of relevant ground literals. The planning task is to compute a sequence of actions $\pi = (a_1, \ldots, a_K)$ that takes the system from $\mathcal{I}$ to a state where $\mathcal{G}$ is satisfied. In our case, we compute plans using the state of the art Fast Downward (FD) system in the \textit{Autotune} configuration~\cite{Helmert_2006}. This heuristic planner adapts its parameters based on instances of the domain under consideration and supports different heuristics and options. We focus on minimizing the \texttt{total cost} $C$ of the plan, i.e., if $c_k^j$ is the cost of action $a_k^j$ in plan $\pi^j$, the objective would be to compute:
\vspace{-0.7em}
\begin{align*}
    \pi^* = \arg \min_{\pi^j} C(\pi^j), \quad C(\pi^j) = \sum_{k=0}^K c_{k}^j
\end{align*}
where the cost of each aciton corresponds to the time taken by the agent to execute it. $\mathcal{P}$ also includes some helper statements that guide this search for plans. Recall that the agent will typically focus on computing a plan for one high-level task at a time, e.g., left panel of Figure~\ref{fig:comparison_plan}; when the agent tries to compute an action sequence to jointly achieve multiple high-level goals, the overall plan length and execution cost can be reduced, e.g., right panel of Figure~\ref{fig:comparison_plan}.

\lstset{basicstyle=\footnotesize\ttfamily}
\captionsetup{font=scriptsize}

\section{Experimental Setup and Results}
We experimentally evaluated four hypotheses:
\vspace{-0.75em}
\begin{enumerate}
\itemsep-10pt
\item[\textbf{H1:}] LLMs are able to accurately anticipate future tasks based on a small number of prompts of task routines.
\item[\textbf{H2:}] LLMs can take into account specific contextual information for task anticipation.
\item[\textbf{H3:}] Considering anticipated tasks as joint goals reduces plan length and plan execution time compared with considering one task at a time.
\item[\textbf{H4:}] Our framework allows the agent to adapt to unexpected successes and failures by interrupting plan execution and replanning if necessary.
\end{enumerate}
We evaluated H1 using three LLMs: PaLM \cite{chowdhery2022palm}, GPT-3.5\cite{brown2020language} and GPT-4\cite{OpenAI2023GPT4TR}. For the other hypotheses we used GPT-4 as the default LLM. For H3, we computed plans using different configurations of the Fast-Downward system~\cite{Helmert_2006}, and we evaluated H4 qualitatively.

\subsection{Experimental Setup}
\label{sec:expt}
We first describe the setup process we followed to experimentally evaluate the hypotheses.

\subsubsection{Prompting LLMs and Planning}
\label{sec:llm-prompting}
We created a dataset $\mathcal{T}$ of high-level tasks in the household environment. These tasks belong to activities such as \textit{Cooking, Cleaning, Washing, Baking}, and \textit{Gardening}. We then generated multiple task routines $\mathcal{R}_i$, each with $\approx 20$ tasks, by sampling tasks from different activities while preserving the relative order of tasks within each activity. 
This process helped us create task routines spanning activities of daily living in a home. 
\begin{figure}[tbp]
\captionsetup{font=scriptsize}
  \includegraphics[width=0.475\textwidth]{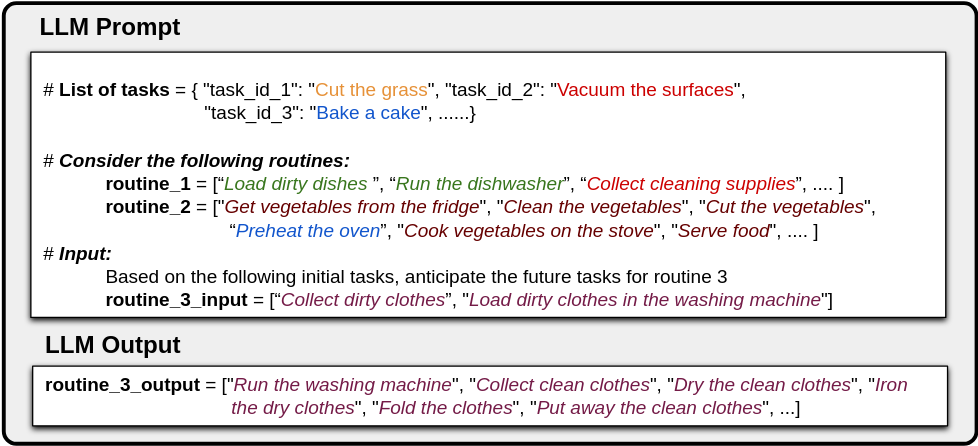}
\setlength{\belowcaptionskip}{-15pt}
\caption{LLM prompting example.}
\label{fig:llm_prompts}
\end{figure}

\begin{table*}[!h]
\centering
\captionsetup{font=scriptsize}
\setlength{\belowcaptionskip}{-7pt}
\begin{tabular}{| >{\centering\arraybackslash} m{1.7cm}|  >{\centering\arraybackslash} m{2.5cm}| >{\centering\arraybackslash} m{3cm}| >{\centering\arraybackslash} m{0.9cm}| >{\centering\arraybackslash} m{2.5cm}| >{\centering\arraybackslash} m{3cm}| >{\centering\arraybackslash} m{0.9cm}|}
\hline
\\[-1em]
 \multirow{2}{*}{LLMs} & \multicolumn{3}{c|}{Without context} & \multicolumn{3}{c|}{With Context} \\
\cline{2-7}
\\[-1em]
 & Miss Ratio (Miss.) $\Downarrow$ & Partial Ordering Count (POC) $\Uparrow$ & $KRCC$ $\Uparrow$ & Miss Ratio (Miss.) $\Downarrow$ & Partial Ordering Count (POC) $\Uparrow$ & $KRCC$ $\Uparrow$  \\
\hline
\\[-1em]
PaLM & 0.361 & 0.974 & 0.993 & 0.034 & 0.994 & 0.996 \\
\hline
\\[-1em]
GPT-3.5-turbo & 0.282 & 0.676 & 0.906 & 0.0698 & 0.806 & 0.976 \\
\hline
\\[-1em]
\textbf{GPT-4} & \textbf{0.037} & \textbf{0.960} & \textbf{0.995} & \textbf{0.0006} & \textbf{1.0} & \textbf{1.0} \\
\hline
\end{tabular}
\caption{Task anticipation performance of LLMs. We ran 500 experiments with $\approx$ 20 tasks per experiment. We observed a significant increase in performance after providing contextual prompts. $\Downarrow$ ($\Uparrow$) implies that lower (higher) values of the corresponding measure represents better performance. Results support \textbf{H1} and \textbf{H2}.}
\label{tabl1}
\end{table*}

We conducted experiments under two configurations. In both configurations, the dataset $\mathcal{T}$ was provided to the LLMs (in JSON format) to minimize hallucinations and to ground our LLM outputs since these outputs are mapped to goal descriptions based on closed set of PDDL statements. Tasks not in $\mathcal{T}$ will be ignored and handling such tasks is beyond the scope of this paper. In the \textbf{without context} configuration, the task routines followed during two individual days were given as input (prompt) to the LLM, which was asked to complete a partially-specified routine comprising two tasks. Figure \ref{fig:llm_prompts} shows an example prompt and LLM output at this stage. In the \textbf{with context} configuration, in addition to the two task routines (as before), we used a method similar to~\cite{silver2023generalized} to provide two contextual examples in form of the partially-specified task inputs and the corresponding expected LLM outputs. The difference between the two configurations was thus the additional contextual prompting provided in the second configuration to guide the LLM. We considered four task anticipation performance measures:
\vspace{-0.5em}
\begin{itemize}
\itemsep-10pt
    \item \textit{Miss Ratio (Miss.)}: ratio of number of tasks \textbf{not} anticipated to the length of the sampled sequence.

\item \textit{Partial Ordering Count (POC)}: measures capability to maintain the relative order of tasks in the routines.

\item \textit{Kendall rank correlation coefficient (KRCC)~\cite{Kendall}}: measures match between predicted and actual task order.
\vspace{-0.5em}
\begin{align*}
    KRCC = \frac{n_c - n_d}{\sqrt{(n_0 - n_1)(n_0 - n_2)}}
\end{align*}
where: \(n_c\) is the number of concordant pairs, \(n_d\) is the number of discordant pairs, \(n_0 = n(n-1)/2\) is the total number of pairs, \(n_1\) and \(n_2\) are the sums of ties in the first and second sequences. A pair of tasks (\(\tau_{1}, \tau_{2}\)), (\(\tau_{3}, \tau_{4}\)) is said to be concordant if \(\text{rank}(\tau_1) > \text{rank}(\tau_3)\) and \(\text{rank}(\tau_2) > \text{rank}(\tau_{4})\) or \(\text{rank}(\tau_1) < \text{rank}(\tau_3)\) and \(\text{rank}(\tau_2) < \text{rank}(\tau_{4})\). If the ranks disagree, the pair is discordant. A tie can occur when there is a repetition of tasks in the anticipated routine, but since we do not consider task repetition, $n_2$ will always be zero.
    \item \textit{Success Ratio}: Fraction of experiments (i.e., prompts) for which tasks were anticipated correctly by the LLM.
\end{itemize}
Recall that to perform any high-level task, the agent has to plan a sequence of finer-granularity actions. In our experiments, the number of actions required to accomplish any given task varied from one to 16, with the initial domain description (for planning) comprising 33 independent actions, five different rooms, 33 objects distributed over 5-10 types, and 19 receptacles. \textit{We thus set up a complex domain for experimental analysis compared with other papers that have explored task anticipation or combined LLMs with PDDL-like representations for planning}. 


\vspace{0.5em}
\subsubsection{Baseline}
\label{sec:baseline}
As a baseline for the task anticipation capability of the LLMs, we sampled 100 routines of tasks and created a probability transition matrix representing the likelihood of transitioning from one task to another within the dataset:
\(
P(\tau_{j}|\tau_{i}) = \frac{Count(\tau_{i}, \tau_{j})}{Count(\tau_{i})}
\)
where $P(\tau_{j}|\tau_{i})$ is the probability of transitioning from task $\tau_{i}$ to task $\tau_{j}$, $Count(\tau_{i}, \tau_{j})$ denotes the number of occurrences of the transition from task $\tau_{i}$ to task $\tau_{j}$ in the dataset, and $Count(\tau_{i})$ represents the total number of occurrences of task $\tau_{i}$ in the dataset. We then created a Markov chain of the tasks such that given an initial task, subsequent tasks are obtained by repeatedly sampling from the probability transition matrix. For the planning experiments, the baseline was planning without considering any anticipated tasks.

\subsection{Experimental Results}
\label{sec:expres}
Next, we describe the results of experimental evaluation.

\subsubsection{Evaluating H1 and H2}
To evaluate \textbf{H1}, we evaluated the ability of PaLM, GPT-3.5 and GPT-4 to anticipate future tasks, as stated in Section~\ref{sec:llm-prompting}. We ran 500 experiments with tasks sampled from our household dataset, with the corresponding results summarized in Table~\ref{tabl1}. Even in the absence of contextual examples, all LLMs maintained the order of tasks in a routine. However, PaLM was not able to correctly anticipate all the tasks; it missed $approx 36\%$ of the tasks. When the LLMs had access to the contextual examples, all three provided very good performance, with GPT-4 providing the correct task order $100\%$ of the time along with a very low Miss Ratio (0.06\%). Overall, the LLMs were able to anticipate tasks based on a limited number of prompts, with performance varying between the three LLMs; all three LLMs provided good task anticipation performance when they had access to the contextual examples. 
These results support hypotheses \textbf{H1} and \textbf{H2}.

Next, we experimented with GPT-3.5-turbo and GPT-4 under specific conditions. Specifically, we arranged tasks from our dataset in a weekly schedule.
After providing the LLM with the routine of tasks during each day of the week, we posed a special prompt that deviated from the expected routine. One example of such a prompt is: "Today is a Monday. I have an urgent meeting in the morning." We observed that
LLMs were able to respond to such prompts and respect the constraint imposed by the prompt while generating the anticipated tasks. For the specific prompt (above), the output included most of the expected tasks for Monday and two extra tasks to set up a laptop and prepare clean clothes right after breakfast. We ran 20 such experiments, and the results are summarised in Table~\ref{h2_llm}. These results further support \textbf{H2}.

\begin{table}[tb]
\centering
\captionsetup{font=scriptsize}
\begin{tabular}{| c | c |}
\hline
 & Success Ratio $\Uparrow$ \\
\hline
\\[-1em]
GPT-3.5-turbo & 0.65 \\
\hline
\\[-1em]
\textbf{GPT-4} & \textbf{0.8} \\
\hline
\end{tabular}
\setlength{\belowcaptionskip}{-5pt}
\caption{Task anticipation of two LLMs with contextual information. Success ratio expressed as a fraction of tasks averaged over 20 experiments. Results support \textbf{H2}.}
\label{h2_llm}
\end{table}
\begin{table}[tb]
\centering
\captionsetup{font=scriptsize}
\begin{tabular}{| >{\centering\arraybackslash} m{1.2cm}|  >{\centering\arraybackslash} m{0.9cm}| >{\centering\arraybackslash} m{0.95cm}| >{\centering\arraybackslash} m{0.7cm}| >{\centering\arraybackslash} m{0.9cm}| >{\centering\arraybackslash} m{1.2cm}| }
\hline
 & Miss. $\Downarrow$ & POC $\Uparrow$ & \tiny{KRCC} $\Uparrow$ & Incorr. $\Downarrow$ & Repeat $\Downarrow$ \\
\hline
\\[-1em]
\textbf{GPT-4} & \textbf{0.0006} & \textbf{1.0} & \textbf{1.0} & \textbf{0} & \textbf{0} \\
\hline
\\[-1em]
Markovian & 0.413 & 0.364 & 0.908 & 6.28 & 1.49 \\
\hline
\end{tabular}
\setlength{\belowcaptionskip}{-15pt}
\caption{Task anticipation of LLMs compared with the Markovian baseline. Values of measures expressed as an average over 500 experiments. Column "Incorr." summarizes the number of anticipated tasks that were incorrect, and "Repeat" shows the average number of repetitions per experiment. Results support \textbf{H1}.}
\label{hist_baseline}
\end{table}

\begin{figure*}[tb]
\centering
\captionsetup{font=scriptsize}
\setlength{\belowcaptionskip}{-12pt}
\setlength{\abovecaptionskip}{5pt}
\includegraphics[height=0.35\textwidth,width=0.78\textwidth,frame]{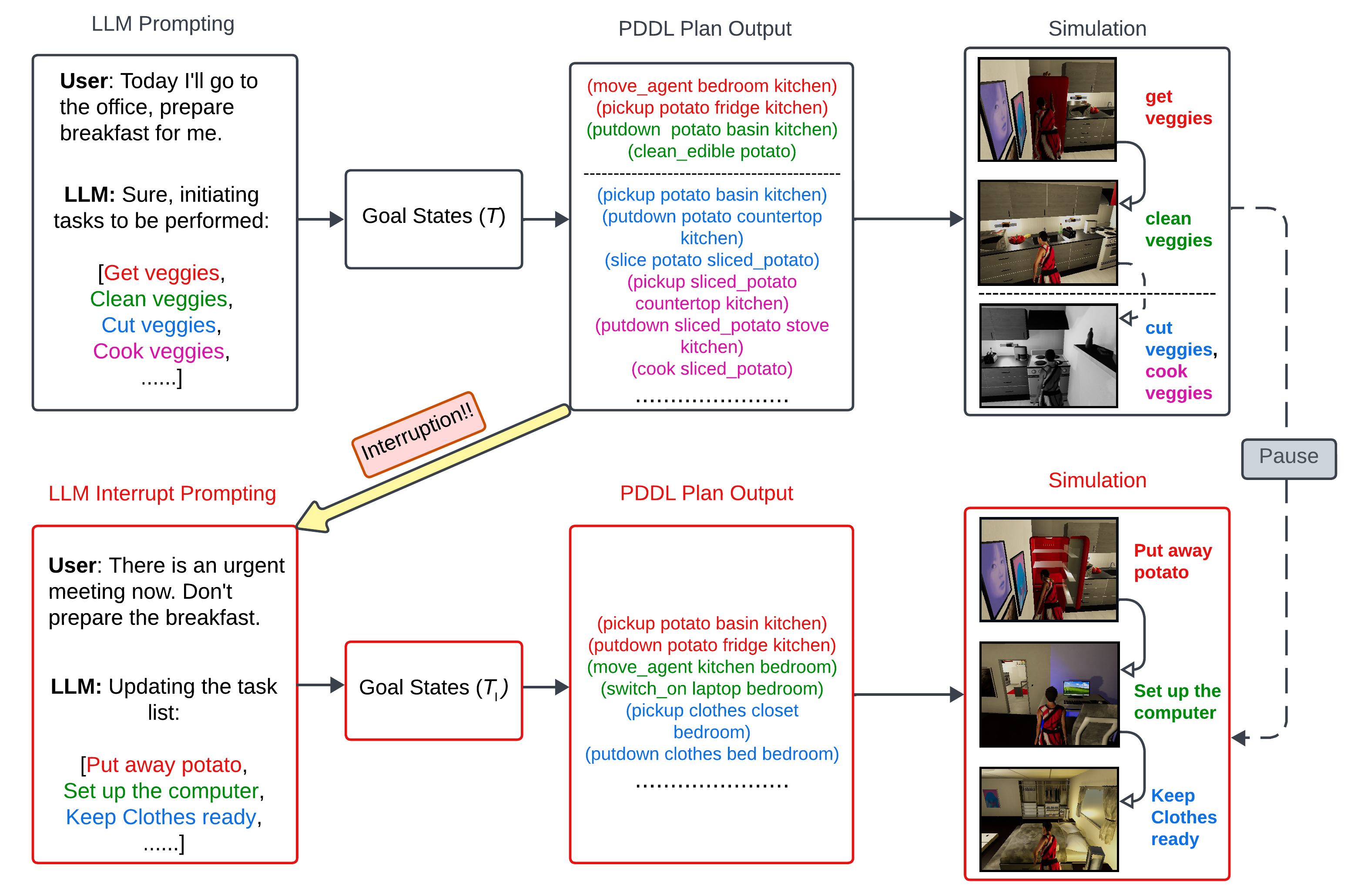}
\caption{An illustrative use case that involved an interruption during the execution of a plan computed to jointly accomplish some anticipated tasks; the agent was able to revise the anticipated tasks and replan as appropriate.}
\label{fig:interrupt}
\end{figure*}
\vspace{-1em}
Next, we compared the task anticipation ability of LLMs with the first-order Markovian baseline described in Section~\ref{sec:baseline}. For the baseline, we generated the routine of tasks based on the transition probability function of the baseline, with the results summarized in Table~\ref{hist_baseline}. Since the baseline was based on a Markov chain, there was no way for the system to recover if it reached a faulty state; it just continued to sample tasks from the faulty state. The deviation of the baseline from the correct routine (and the LLM's output) is shown in the columns labeled \textit{Incorr.} and \textit{Repeat} in Table~\ref{hist_baseline}. Unlike this baseline, the LLMs were able to 
learn from limited prompts representing domain-specific preferences. These results further support \textbf{H1}.

\subsubsection{Evaluating hypothesis H3}
To measure the impact of considering the anticipated tasks during planning, we measured the cost of executing the computed sequence of finer-granularity actions. Specifically, we prompted the LLM with contextual examples and asked it to provide different number of anticipated tasks for a partially-specified routine. These anticipated tasks were considered as joint goals by the planning system, with the resultant plan of actions being executed by the agent. Since the cost of the actions (in the domain description) was based on the execution time, we used the total cost of any executed plan as the execution time (in seconds) of the plan. The results of these experiments are summarized in Table~\ref{table:cost}. 

In these experiments, we varied the number of anticipated tasks from zero (``Myopic"), with the agent planning to perform one task at a time, to six, i.e., with six tasks anticipated and considered jointly during planning. As the number of anticipated tasks increased, the search time limit provided to the planner was increased by units of 30 seconds. For example, when the agent jointly planned an action sequence for the current task and the next (anticipated) task, the search time limit was set to 60 seconds. In addition, we organized these experiments as paired trials with the initial conditions and the overall sequence of assigned tasks being the same in each paired trial across the different number of anticipated tasks. Each value reported in Table~\ref{table:cost} is the average of 10 repetitions of the corresponding experiment.


\begin{table}[tb]
\centering
\begin{tabular}{|>{\centering\arraybackslash} m{1.3cm}|>{\centering\arraybackslash} m{1.8cm}|>{\centering\arraybackslash} m{1.1cm}|>{\centering\arraybackslash} m{1cm}|>{\centering\arraybackslash} m{1cm}|}
\hline
 \multicolumn{2}{|c|}{Number of tasks anticipated $\rightarrow$} & \multirow{2}{*}{0(Myopic)} & \multirow{2}{*}{3 (Ours)} & \multirow{2}{*}{6 (Ours)} \\
\cline{1-2}
 Planner & Parameters & & & \\
\cline{1-5}
\\[-1em]
\multirow{2}{*}{\textbf{AT-1}} & Plan Length & 70.3 & 65.2 & 61.8 \\
\cline{2-5}
\\[-1em]
 & Exe. Time & 2051 & 1658 & 1390 \\
\cline{1-5}
\\[-1em]
\multirow{2}{*}{LAMA} & Plan Length & 65.7 & 62.5 & 61.2 \\
\cline{2-5}
\\[-1em]
 & Exe. Time & 1835 & 1613 & 1599 \\
\cline{1-5}
\\[-1em]
\multirow{2}{*}{AT-2} & Plan Length & 67.2 & 64.3 & 60.2 \\
\cline{2-5}
 & Exe. Time & 1847 & 1591 & 1377 \\
\hline
\end{tabular}
\setlength{\belowcaptionskip}{-15pt}
\caption{Planning and execution performance with three different values of the number of anticipated tasks, based on three different configurations of the Fast Downward planner; AT1, AT2, and LAMA correspond to configurations \textit{seq-sat-fd-autotune-1}, \textit{seq-sat-fd-autotune-2}, and \textit{seq-sat-lama-2011} respectively. \textit{Exe. Time} denotes the plan execution time in seconds. Results support hypothesis \textbf{H3}.}
\label{table:cost}
\end{table}

Table~\ref{table:cost} shows the plan length and execution time as the number of anticipated tasks changes from zero to three and six, under three configuration options supported by FD: \textit{seq-sat-fd-autotune-1, seq-sat-lama-2011}, and \textit{seq-sat-fd-autotune-2}. For each configuration, as the number of anticipated tasks increased, the plan length decreased. There may be positive interaction between goals, with actions executed to achieve one goal leading to conditions that are essential preconditions for actions to be executed to achieve a subsequent goal. In such cases, planning jointly for the two goals will not make a big difference compared with pursuing one goal at a time. However, positive interaction between goals was unlikely in our experiments because tasks were anticipated by the LLM at a high level of abstraction.
The observed reduction in plan length in Table~\ref{table:cost} thus indicates that anticipating and planning to jointly achieve multiple tasks enables the agent to complete all the tasks by executing fewer actions compared with not considering anticipated tasks. Furthermore, the plan execution time decreased with an increase in the number of anticipated tasks, indicating that the agent became more efficient when it interleaved the actions for different tasks. In these planning trials, the agent often had to use the available planning time limit to compute the plans, particularly when the task required it to sequence multiple actions and when multiple anticipated tasks had to be considered. As a result, although there was a reduction in plan length and execution time, the planning time reached a plateau and did not change much during the experiments summarized in Table~\ref{table:cost}. These results support \textbf{H3}.

\begin{figure}
    \centering
    \footnotesize{
        \includegraphics[height=0.3  \textwidth,width=\linewidth]{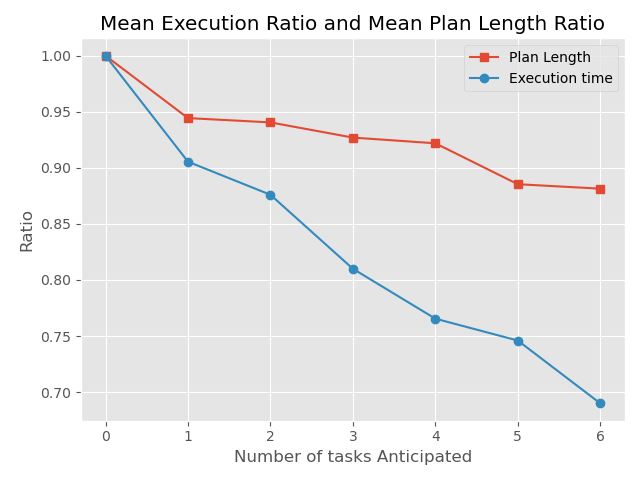}
    
    \setlength{\belowcaptionskip}{-15pt}
    \caption{Plan execution time and plan length with the number of anticipated tasks ranging from 1-6. Values expressed as ratio of the corresponding values in the myopic case (x=0). We observed a reduction of 31\% in execution cost, and 12\% in plan length.}
    \label{fig:trend}
    }
\end{figure}

\vspace{-1em}
Figure~\ref{fig:trend} further illustrates the results of these planning experiments for the \textit{seq-sat-fd-autotune-1} configuration. Since the initial state and set of tasks varied between each set of paired trials, averaging the numbers (e.g., for execution time) across these trials may not be meaningful. In each paired trial, we thus expressed the execution time and plan length of each instance that involved one or more anticipated tasks, as a fraction of the corresponding execution time and plan length (respectively) obtained without any task anticipation (i.e., "myopic"). The average of these fractions over 10 repetitions is shown in Figure~\ref{fig:trend}. We observed a substantial ($\approx 31\%$) reduction in execution time and plan length ($\approx 12\%$) as the number of anticipated tasks increased to six. We thus conclude that anticipating future tasks and considering them during planning leads to more efficient performance; these results further support \textbf{H3}.

\subsubsection{Evaluating hypothesis H4}
Finally, we qualitatively evaluated H4. In these experiments, the execution of a plan generated by a planner was randomly interrupted by the user. Each interrupt was accompanied by a prompt by the user, leading to a change in the agent's operation. In particular, the state of the environment before the interrupt and the prompt (used for interrupting) were sent to the LLM to generate a new routine of tasks to be accomplished by the agent. While some actions in our domain are irreversible, e.g., we cannot put the pieces of a cut fruit back together, our combination of LLMs and action planning was able to undo the effects of some actions when that is appropriate. For example, in Figure~\ref{fig:interrupt}, action execution was interrupted by the prompt ``There is an urgent meeting now. Do not prepare breakfast". In this situation, the agent was able to plan suitable actions and put the vegetables back in their original location because it was no longer necessary to cook a hot breakfast; the agent instead computed action sequences to prepare the laptop and suitable clothes for the meeting.

\section{Conclusion and Future Work}
This paper described a framework for task anticipation and action execution by an agent in complex household environments. The framework leverages the generic knowledge encoded in LLMs for high-level task anticipation based on limited prompts, and plans a sequence of finer-granularity actions based on the domain-specific knowledge encoded in PDDL to jointly accomplish the anticipated tasks. We demonstrate a substantial reduction in action execution time and plan length in comparison with a planning system that does not consider anticipated actions. In future work, we intend to explore scalability of this framework to more complex domains, incorporate probabilistic planning, and explore implementation on a physical robot assistant. We will also explore the ability of LLMs to automatically adapt task anticipation to the preferences and needs of specific individuals in the household in which the agent is operating.

\bibliographystyle{IEEEtran}

\balance
\end{document}